\documentclass{article} 
\usepackage{NORS,times}

\usepackage{amsmath,amsfonts,bm}

\def\eqref#1{equation~\ref{#1}}

\def\1{\bm{1}}

\DeclareMathAlphabet{\mathsfit}{\encodingdefault}{\sfdefault}{m}{sl}
\SetMathAlphabet{\mathsfit}{bold}{\encodingdefault}{\sfdefault}{bx}{n}

\usepackage{hyperref}
\usepackage{url}
\usepackage[pdftex]{graphicx}
\usepackage{multirow}
\usepackage{amsmath}
\usepackage{amsthm}

\title{Neural Operator with Regularity Structure for Modeling Dynamics Driven by SPDEs}

\author{Peiyan Hu$^{1}$\thanks{This work was done when the first author was visiting Microsoft Research Asia.}, Qi Meng$^{2}$\thanks{Corresponding E-mail: meq@microsoft.com.} , Bingguang Chen$^{3}$, Shiqi Gong$^{3}$, Yue Wang$^{2}$, Wei Chen$^{3}$,\\
\textbf{Rongchan Zhu$^{4}$, Zhi-Ming Ma$^{3}$, Tie-Yan Liu$^{2}$}\\
$^{1}$University of Chinese Academy of Sciences,
$^{2}$Microsoft Research Asia\\
$^{3}$Chinese Academy of Sciences,
$^{4}$Bielefeld University
}

\iclrfinalcopy 
\begin{document}

\maketitle

\begin{abstract}
Stochastic partial differential equations (SPDEs) are powerful mathematical models for modeling dynamics in many areas including atmospheric sciences and physics.
Neural Operators are deep learning based approach which are proposed for solving parametric PDEs. However, the existing neural operators does not take the SPDEs into consideration, which usually have poor regularity \footnote{Roughly speaking, regularity describes the smoothness of a function.} due to the driving noise. As the theory of regularity structure has achieved great successes in analyzing SPDEs and provides the concept \emph{model feature vectors} that well-approximate SPDEs' solutions, we propose the Neural Operator with Regularity Structure (NORS) which incorporates the feature vectors for modeling dynamics driven by SPDEs. We conduct experiments on various SPDEs including the dynamic $\Phi^4_1$ model and the 2d stochastic Navier-Stokes equation, and the results demonstrate that the NORS is efficient and achieves one order of magnitude lower error with a modest amount of data.
\end{abstract}

\section{Introduction}

Stochastic partial differential equations (SPDEs), which generalizes PDE via random force terms and coefficients, are significant tools for modeling dynamics in many areas including atmospheric sciences \citep{atmospheric}, physics \citep{physics}, biology \citep{biology}, economics \citep{econoics}, etc. SPDEs are used to study statistical mechanics of the dynamics systems, e.g., stochastic Navier-Stokes equations models the statistics of turbulent flows \citep{buckmaster} in atmospheric science and the $\Phi^4$ model arises in the stochastic quantisation of quantum field theory \citep{phi43}. 
Since SPDEs relate to many scientific open problems, studying the solution of SPDEs from both mathematical proving and numerical methods is a hot research direction in both math and physics. 

Inspired by recent advances in using AI techniques to accelerate scientific computing, we study using deep learning methods for modeling the solution of SPDEs. There have been deep learning models arising for modeling dynamics governed by PDEs such as Neural Operators \citep{Kovachki} (which model the map between infinite-dimensional functions and agree with the case of learning solutions of a family of parametric PDEs), DeepONet \citep{lu2019deeponet}. However, SPDEs usually have poor regularity w.r.t the time variable for function-valued noise and singularity w.r.t space for space-time white noise that these models do not take into consideration. Thus, how to properly encoding the SPDEs' information and represent the solutions of the SPDEs are deserved to be investigated. 

To deal with the singularity of SPDEs, we incorporate the feature engineering with regularity proposed by \cite{chevyrev} with neural operator to deal with the regularity problem. The feature engineering with regularity is to project the driving noise and initial conditions to the \emph{model feature vectors}, which composes a basis of SPDEs' mild solution in regularity structure theory. According to the Schauder estimates for related linear operator, the model feature vectors have improved regularity. 

\textbf{Our Contributions} We introduce the Neural Operator with Regularity Structure (NORS) that extends the Neural Operators. This deep-learning-based method has three advantages as follows: (1) The NORS can solve equations with changing driving force, which is beyond the Neural Operators' capability because of its requirement of regularity. The detailed theory about regularity structure is provided in Section \ref{Preliminary}.
(2) The NORS utilizes more information from the equations themselves as we take models as our features, which contain the information of the SPDEs' differential operators and then reduce the sample complexity and leads to lower loss.
(3) Experiments show that NORS inherits resolution-invariant property of Fourier Neural Operator (FNO), which can stay accurate across different resolutions. (4) We test the NORS on the dynamic $\Phi^4_1$ model, reaction-diffusion equation with linear multiplicative noise and the 2d stochastic Navier-Stokes equation. Using the NORS, both the testing accuracy and sample complexity are enhanced. Specifically, the error is one order of magnitude lower than other baselines.

\section{Related work}
There have been several popular deep-learning-based methods for modeling the solution of parametric partial differential equations \citep{lu2019deeponet, patel2021physics, Kovachki, Li, bhattacharya2020model, nelsen2021random, li2020multipole}. 
For example, the Neural Operator \citep{Kovachki} and Fourier Neural Operator (FNO) \citep{Li} are representatives which are mesh-independent model, whose architectures are inspired by Green's functions of PDEs. 
Since the solution of SPDEs is determined by both the initial condition and the force, capturing the structure of the force (e.g., the space-time white noise) is beyond the capability of these models. To handle the case that SPDEs' solutions depend simultaneously on the initial condition $u_0$ and the force term $\xi$, \citet{Salvi} introduce the neural stochastic partial differential equation (Neural SPDE), which parameterizes the kernels according to Duhamal's fix-point formula for SPDE whose linear differential operators can generate semigroups.  
In this paper, we adopt another way which first projects the initial condition and force to a set of models, {which follows \citep{chevyrev}. \citep{chevyrev} explored the feature engineering with regularity structure but was limited to the linear regression, while the NORS combines the features with neural operator, results in a more strong tool with the regularity structure.} One main difference between NORS and Neural SPDE \cite{Salvi} is that NORS assumes known explicit form of the linear part (i.e., the semigroup of the SPDE) and leverage it to generate model feature vectors, while Neural SPDE does not leverage the information and regards both linear and non-linear parts of the SPDE as black-box models. Since the \emph{model feature vectors} incorporate more prior (including the kernel, the initial condition, and the force) of the SPDE, it is expected to have a better generalization and lower sample complexity.

\section{Preliminary}
\label{Preliminary}

In this section, we introduce background on the regularity structure theory \citep{Hairer} of SPDEs. Consider an SPDE on $[0,T]\times D$ with the following form
\begin{align}
    \partial_tu-\mathcal{L}u = \mu(u,\partial_1u,\cdots &,\partial_du)+\sigma(u,\partial_1u,\cdots,\partial_du)\xi,\nonumber\\
    & u(0,x)=u_0(x), \label{SPDE}
    \end{align}

where $x\in D\subset\mathbb{R}^d$, $t\in[0,T]$, $\mathcal{L}$ is a linear differential operator, $\xi$ is the space-time white noise, $u_0:D\rightarrow\mathbb{R}$ is the initial condition. Under local Lipschitz condition on $\mu,\sigma$ with respect to suitable norm, this SPDE has a unique mild solution \citep{Hairer,Salvi}:
{\small\begin{equation} 
u_t=e^{t\mathcal{L}}u_0+\int_0^t e^{(t-s)\mathcal{L}}\mu(u_s,\partial_1u_s,\cdots ,\partial_du_s) ds +\int_0^t  e^{(t-s)\mathcal{L}}\sigma(u_s,\partial_1u_s,\cdots,\partial_du_s)\xi ds. \label{mild}
\end{equation}}Thus in the field of SPDEs, the solution is determined by both the initial condition and the force term. The design of deep learning models such as Neural Operators does not consider the solution structure of SPDEs, {which should involve the low-regularity driving force term. Therefore, these models are not universal approximators for SPDEs.}


It is then natural to utilize the regularity structure theory to help handle the regularity problem. The concept \emph{model} in the regularity structure is a collection of \emph{model feature vectors}, which are multi-dimensional signals designed to approximate solutions of SPDEs even with low regularity regimes. The motivation comes from Picard theorem and Taylor expansion. According to the representation of the mild solution in Eqn.(\ref{mild}), we define two linear operators $I[f](t)=\int_{0}^te^{(t-s)\mathcal{L}} f(s)ds$ and $I_c[u_0](t)=e^{t\mathcal{L}}u_0$ 
for any function $f$ defined on $[0,T]\times D$ to $\mathbb{R}^d$. 
Picard theorem shows that the following recursive sequence approximates the solution $u$ of equation (\ref{SPDE}) as $n\rightarrow\infty$
{\small\begin{equation}
    \begin{aligned}
    &u^0_t = I_c[u_0]_t,\quad
    u^{n+1}_t = I_c[u_0]_t+I[\mu(u^n)+\sigma(u^n\xi)]_t. 
    \end{aligned} \label{iteration}
\end{equation}}
Using Taylor expansion, we then have the recursive sequence that can approximate $u$ as $m,l,n\rightarrow\infty$
{\small\begin{equation}
    \begin{aligned}
    &u^{0,m,l}_t = I_c[u_0]_t,\\
    &u^{n+1,m,l}_t = I_c[u_0]_t + \sum^m_{k=0}\frac{\mu^{(k)}(0)}{k!}I[(u^{n,m,l})^k]_t + \sum^l_{k=0}\frac{\sigma^{(k)}(0)}{k!}I[(u^{n,m,l})^k\xi]_t. \label{feature}
    \end{aligned}
\end{equation}}Then, the solution of SPDE can be approximated by weighted sum of the features $I[(u^{n,m,l})^k], I[(u^{n,m,l})^k], l=0,\cdots,k; m=0,\cdots,k$, where we call $n$ as the height and $m, l$ as the width in the approximation. Motivated by this, \citet{chevyrev} develops tool for feature engineering of SPDEs. By the regularity structure theory, the model feature vectors are obtained by integrals of functionals of $u_0$ and $\xi$ (as $I$ and $I_c$ are convolution operations), whose regularity is proved to be better due to the polishing effect of integrals \citep{Salvi}. To avoid the number of model feature vectors grows exponentially, the \emph{height} of the features is constrained according to the regularity of the SPDE. Please refer the details about the generation of model feature vectors and its degree constraints in Appendix A.1

\begin{figure}[t]
\begin{center}
\centerline{\includegraphics[width=1\linewidth]{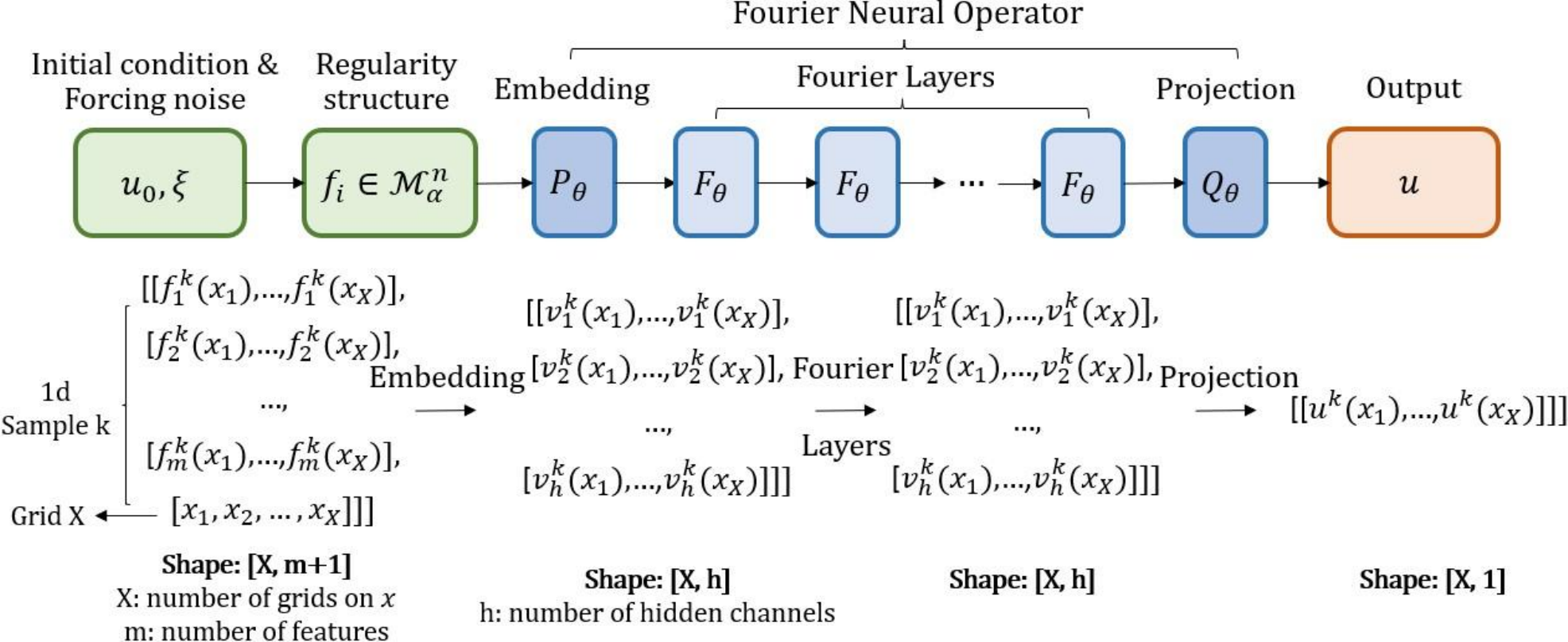}}
\end{center}
\caption{The architecture of our model and the shape of the data of one sample in the 1d case.}
\label{architecture}
\end{figure}
\section{Learning SPDE Solution via model feature vectors }
We move on to introducing the Neural Operator with Regularity Structure (abbrev. NORS).  For given SPDE which has the form in Eqn.(\ref{SPDE}), our goal is to learn its solution solution $u_T$ at given time point $T$ under initial condition $u_0$ which is assumed to be generated by a parametric distribution.  
According to Eqn. (\ref{iteration}) and (\ref{feature}), {the solution depends continuously on the model feature vectors not on the initial condition $u_0$ and noise $\xi$. Therefore, NORS first maps $u_0$ and $\xi$ to the model feature vectors,} and then we use Fourier Neural Operator (abbrev. FNO) to learn the continuous map from the model feature vectors to the solution.  

To represent the continuous input functions $u_0$ and $\xi$, we discretize the space-time domain $D\times[0, T]$ with $D\subset \mathbb{R}^d$ onto the grid $O_{X_1}\times\cdots\times O_{X_d}\times O_T$. Then we use the values of the continuous function on the grid points to represent them. For one sample of $u_0$ and $\xi$, we first get the model feature vectors $\mathcal{M}$ of $(u_0,\xi)$ according to data and the form of the equation. As $\mathcal{M}=\{f_i\}_{i=1,\cdots,m}$ are a set of continuous functions, we also use its value on discrete grids to represent them. By concatenating all the model feature vectors $f_i$ and the grid $O_{X_1}\times\cdots\times O_{X_d}$, we get the inputs $w^0$. Compared with original FNO, the number of input channels are enlarged which contains both the original grids and the model feature vectors. 

Then, $w^0$ is fed into the FNO and the forward process is expressed as {\begin{equation}
    \begin{aligned}
    v^0(x) = P_{\theta_{in}}(w^0(x)); \quad
    v^{i+1}(x) = F_{\theta_i}(v^i(x)); \quad
    \hat{u}_T(x) = Q_{\theta_{out}}(v^K(x))
    \end{aligned}
\end{equation}}
for any $x\in D$, where  $\theta_{in}$, $\theta_{out}$, $\theta_i, i=0,\cdots,K-1$ are learnable weights, $P_{\theta_{in}}:\mathbb{R}^{X_1\cdots\times X_d\times(m+d)}\rightarrow\mathbb{R}^{X_1\cdots\times X_d\times h}$ is an embedding neural network to project the input to the latent feature space, $F_{\theta_i}:\mathbb{R}^{X_1\cdots\times X_d\times h}\rightarrow\mathbb{R}^{X_1\cdots\times X_d\times h}$ is a Fourier layer \citep{Li} which approximates the iteration in Picard's iteration, $Q_{\theta_{out}}:\mathbb{R}^{ X_1\cdots\times X_d\times h}\rightarrow\mathbb{R}^{ X_1\cdots\times X_d\times d}$ is a embedding layer to project the latent feature to the output. Here, $X_i$ is the number of grids on dimension $x_i$, $h$ is the number of hidden channels, and $m$ (the number of model feature vectors) and $d$ (the dimension of region $D$) are defined before. Since only the FNO contains trainable weights, defining loss function between $\hat{u}_T(x)$ and the groundtruth $u(x)$ can guide the optimization to learn the weights of FNO. A demonstration of our model 
is shown in Figure \ref{architecture}.

\section{Experiments}
\label{experiment}

We compare the NORS with other baselines on some significant equations. We adopt similar tasks as \cite{Salvi}, which include the \emph{l2} error on two settings: in the setting ($\xi\mapsto u$), the noise $\xi$ changes while the initial condition $u_0$ is fixed; in the setting ($(u_0,\xi)\mapsto u$), both $\xi$ and $u_0$ vary across samples.  
We note that in the following equations, we only consider periodic boundary conditions, but Dirichlet or Neumann boundary conditions can also be easily complemented. To save space, the details about the construction of the model $\mathcal{M}$ of each SPDE are put into Appendix A.1. 
To utilize the NORS, the assumption is that the form of the differential operator $\mathcal{L}$ is known, which is different from FNO and NSPDE.
We use 32 hidden channels and 4 Fourier layers for our NORS in all experiments.
We use the Adam optimizer to train for 500 epochs with an initial learning rate of 0.001 in the first two experiments and the 2d stochastic Navier-Stokes equation on $64\times64$ grid, and 0.01 for the 2d stochastic Navier-Stokes equation on $16\times16$ grid (after grid search) that are halved every 100 epochs. We randomly split the dataset into training and test sets by 5:1. The NORS codes are deposited in GitHub at \url{https://github.com/Peiyannn/Neural-Operator-with-Regularity-Structure.git}. %

\subsection{Dynamic $\Phi^4_1$ Model}

We first consider the dynamic $\Phi^4_1$ model with the periodic boundary condition. It takes the form
{\begin{equation}
    \begin{aligned}
    \partial_t u-\Delta u = 3u-u^3+\sigma\xi,\quad(t,x)\in[0,0.05]\times[0,1]\\
    u(t,0) = u(t,1),\quad\rm(Periodic\ BC)\\
    u_0(x) = u(0,x) = x(1-x) + \kappa\eta(x),
    \end{aligned}
\end{equation}}
where $\xi$ is the space-time white noise scaled by $\sigma = 0.1$, $\eta(x) = \sum^{k=10}_{k=-10}\frac{a_k}{1+|k|^2}sin(\lambda^{-1}k\pi(x-0.5)),\quad\rm with\ a_k\sim\mathcal{N}(0,1)$ with
 $\lambda=2$, and $\kappa=0$ or $0.1$ corresponding to the initial condition is fixed or not. The setting follows \citep{Salvi}, while we generate the data as \citep{chevyrev}.
 
For this equation, the differential operator $\mathcal{L}$ is $\Delta$, according to which the operator $I$ and $I_c$ of the model $\mathcal{M}^n$ is given by $I[f](t)=\int_{0}^te^{(t-s)\Delta} f(s)ds$ and $I_c[u_0](t)=e^{t\Delta}u_0$, where $\Delta$ is the Laplace operator on $D$ and 
$f:[0,T]\times D\rightarrow\mathbb{R}$.

The result is shown in Table \ref{table:phi41}. We consider two settings, in both of which our architecture outperforms other benchmarks a lot. Even compared with the lowest error of all baselines, our result is about a tenth of it in the $(u_0,\xi)\mapsto u$ setting, while the result of $\xi\mapsto u$ setting is even better. We also note that our model can perform well with few data and low height.

\begin{table}[t]
\caption{\textbf{Dynamic $\Phi^4_1$ model}. We consider the $l2$ error of the baselines and our model($n=2$ or $3$) in two settings with training data size $N=1000$ or $10000$.}
\label{table:phi41}
\begin{center}
\begin{tabular}{lcc|cc}
\multirow{2}{*}{\bf Model} & \multicolumn{2}{c}{$N=1000$} & \multicolumn{2}{c}{$N=10000$} \\
\cline{2-5}
& $\xi\mapsto u$ & $(u_0,\xi)\mapsto u$ & $\xi\mapsto u$ & $(u_0,\xi)\mapsto u$
\\ \hline 
FNO      & 0.013 & 0.030 & 0.003 & 0.024 \\
NSPDE    & 0.044 & 0.042 & 0.024 & 0.039 \\
\cline{1-5}
Ours($n=3$) & \textbf{0.0003} & \textbf{0.0011} & \textbf{0.0002} & \textbf{0.0004} \\
Ours($n=2$) & \textbf{0.0002} & \textbf{0.0012} & \textbf{0.0002} & \textbf{0.0004} \\
\end{tabular}
\end{center}
\end{table}

\subsection{Reaction-Diffusion Equation with Linear Multiplicative Forcing}

As the dynamic $\Phi^4_1$ model is a parabolic equation with additive forcing, we then consider a parabolic equation with multiplicative forcing, which is given by
{\begin{equation}
    \begin{aligned}
    \partial_t u-\Delta u &= 3u-u^3+\sigma u \xi,\quad(t,x)\in[0,0.05]\times[0,1]\\
    u(t,0) &= u(t,1),\quad\rm(Periodic\ BC)\\
    u_0(x) &= u(0,x) = x(1-x) + \kappa\eta(x),
    \end{aligned}
\end{equation}}where $\xi$ is the space-time white noise scaled by $\sigma = 0.1$, and $\eta(x)$ is the same as the $\Phi_1^4$ model. The generation of data is the same as the way in \citep{chevyrev}. As the form of the operator $I$ and $I_c$ of this equation is the same as the $\Phi^4_1$ model, the model $\mathcal{M}^n$ can be constructed similarly. Please check the details in Appendix A.1.

The results in Table \ref{table:multi} show that our model has one order of magnitude lower error in both of the two settings. The experiments on the two equations clearly show that the effectiveness of the \emph{model feature vectors} and the worse generalization of FNO on SPDEs.

\begin{table}[t]
\caption{\textbf{Reaction-Diffusion equation with linear multiplicative forcing}. We compare the $l2$ error of the baselines and our model($n=2$ or $3$) with training data size $N=1000$ or $10000$.}
\label{table:multi}
\begin{center}
\begin{tabular}{lcc|cc}
\multirow{2}{*}{\bf Model} & \multicolumn{2}{c}{\bf $N=1000$} & \multicolumn{2}{c}{\bf $N=10000$} \\
\cline{2-5}
& $\xi\mapsto u$ & $(u_0,\xi)\mapsto u$ & $\xi\mapsto u$ & $(u_0,\xi)\mapsto u$
\\ \hline 
FNO      & 0.0036 & 0.0063 & 0.0035 & 0.0037 \\
NSPDE    & 0.0016 & 0.0062 & 0.0012 & 0.0026 \\
\cline{1-5}
Ours($n=3$) & \textbf{0.0006} & \textbf{0.0005} & \textbf{0.0005} & \textbf{0.0003} \\
Ours($n=2$) & \textbf{0.0006} & \textbf{0.0006} & \textbf{0.0005} & \textbf{0.0003} \\
\end{tabular}
\end{center}
\end{table}

\subsection{2d Stochastic Navier-Stokes Equation}

As both NSPDE and our model claim well performance with different resolutions, we evaluate both this property of NSPDE and our model on a 2d Navier-Stokes equation for an incompressible flow:
\begin{equation}
    \partial_t w-\nu\Delta w = -u\cdot\nabla w+f+\sigma\xi,\quad(t,x)\in[0,0.05]\times[0,1]^2  
    \label{equa:NS}
\end{equation}
\begin{equation}
    \omega(0,x) = \omega_0(x)
\end{equation}
where $u$ is the velocity field, $\omega=\nabla\times u$ is the vorticity, $\omega_0$ is the initial vorticity, $f$ is the deterministic force defined as in \citep{Li}, $\xi$ is the random force rescaled by $\sigma=0.05$ defined as in \citep{Salvi}, and the viscosity parameter $\nu=10^{-4}$.  The generation of space-time white noise follows \citep{Salvi} and we adopt the the data generator in \citep{Li} by replacing the force term to generate the ground-truth for training. 

Our target is to model the vorticity $\omega$, which is harder to learn compared with the velocity $u$. According to the form of the equation,  the operation $I$ and $I_c$ in model $\mathcal{M}^{n}$ is defined as $I[f](t)=\int_{0}^te^{(t-s)\nu\Delta} f(s)ds$ and $I_c[\omega_0](t)=e^{t\nu\Delta}\omega_0$, where $\Delta$ is the Laplace operator defined on the 2d space.
While solving the 2d Navier-Stokes equation on $64\times64$ grid, we train the model on $64\times64$ and $16\times16$ grid respectively to test the property mentioned at the beginning of this subsection.

As shown in Table \ref{table:NS}, the error of our model is one order of magnitude lower in the $\xi\mapsto \omega$ and $(\omega_0,\xi)\mapsto \omega$ settings. Besides, we solve the equation on $64\times64$ grid, then train on the $64\times64$ grid and $16\times16$ grid. From the results, we verify that our model can keep accurate when the resolution changes.

\begin{table}[t]
\caption{\textbf{2d stochastic Navier-Stokes equation}. We compare the $l2$ error of the baselines and our model($n=2$ or $3$) in two settings with 1000 training samples. While solving the equation on $64\times64$ grid, we train the model on $64\times64$ and $16\times16$ grid respectively.}
\label{table:NS}
\begin{center}
\begin{tabular}{lcc|cc}
\multirow{2}{*}{\bf Model} & \multicolumn{2}{c}{$64\times64$ grid} & \multicolumn{2}{c}{$16\times16$ grid} \\
\cline{2-5}
& $\xi\mapsto \omega$ & $(\omega_0,\xi)\mapsto \omega$ & $\xi\mapsto \omega$ & $(\omega_0,\xi)\mapsto \omega$ 
\\ \hline 
NSPDE    & 0.039 & 0.031 & 0.074 & 0.063 \\
Ours($n=3$) & \textbf{0.0017} & \textbf{0.0029} & \textbf{0.0020} & \textbf{0.0034} \\
Ours($n=2$) & \textbf{0.0018} & \textbf{0.0028} & \textbf{0.0022} & \textbf{0.0030} \\
\end{tabular}
\end{center}
\end{table}


\section{Conclusion and Future Work}

In this work, we introduce NORS as a strong SPDE-solving tool with the zero-shot property. By incorporating the regularity structure, the NORS absorbs both the advantages of Neural Operators and regularity structure, and makes up for the shortcomings. Not only can the NORS capture the stucture of noise with low regularity, but also has a much lower error. In the future, as the NORS requires that the differential operator $\mathcal{L}$ is already known, we can extend this method by parameterizing the kernel, which will be able to handle the inverse problem that some part of the equations is unknown. 


\bibliography{NORS}
\bibliographystyle{NORS}

\appendix
\section{Appendix}

\subsection{Model feature vectors}
We review the method to generate the \emph{model feature vectors} introduced in \cite{chevyrev}.  The two types of initial signals are the \emph{forcing} $\xi$ and  functions $\{u^i\}_{i\in\mathcal{J}}$ derived from initial conditions, where $\mathcal{J}$ is the initial index set. Usually, $u^i=I_i[u_0]$, where $I_i$ is a linear operator determined by the specific form of equations. For the SPDE in Eqn.(1) in the main paper, $\mathcal{J}={c}$ and $u^c=I_c[u_0]$.

Fix the \emph{height} $n\in\mathbb{N}$ and the coefficient $\alpha=(m,l,p,q)\in\mathbb{N}^4$ . Then the \emph{model} $\mathcal{M}^n_{\alpha}$ of $(u^i,\xi)$ is defined inductively by
\begin{align}
    \mathcal{M}^0_{\alpha} = \{u^i\}_{i\in\mathcal{J}},
\end{align}
\begin{equation}
    \begin{aligned}
    \mathcal{M}^n_{\alpha} = \{I[\xi^j\prod^k_{i=1}\partial^{\mathbf{a}}f]: f\in\mathcal{M}^{n-1}_{\alpha}, \mathbf{a}\in\mathbb{N}^d,|\mathbf{a}|\le q, j,k\in\mathbb{N}, 0\le j\le p,\\
    1\le k+j\le m \textbf{1}_{j=0}+l\textbf{1}_{j>0}\}\cup\mathcal{M}^{n-1}_{\alpha},
    \end{aligned}
\end{equation}

where $\mathbf{a}=(a_1,...,a_d), \partial^{\mathbf{a}}=\partial_1^{a_1}\cdots\partial_d^{a_d}=\frac{\partial^{a_1}}{\partial x^{a_1}_1}\cdots\frac{\partial^{a_d}}{\partial x^{a_d}_d}, |\mathbf{a}|=\sum^d_{i=1}a_i$, $m$ is the \emph{additive width}, $l$ is the \emph{multiplicative width}, $p$ is the \emph{forcing order} and $q$ is the \emph{differentiation order}. The more specific application differs across SPDEs, and is provided in the following.

To avoid the number of model feature vectors grows exponentially, we constraint it with the \emph{degree} function, that is, only elements do not exceed a certain degree will be involved. The degree deg$:\mathcal{M}^n_{\alpha}\rightarrow\mathbb{R}$ satisfies
\begin{align}\label{equa:deg}
    \rm deg\it I[f]=\beta+\rm deg\it f,\quad \rm deg\it \partial^{a_i}f=\rm deg\it f-|a_i|,\quad \rm deg\it \prod^k_{i=1}f=\prod^k_{i=1}\rm deg \it f,
\end{align}
where $\beta$ is up to the operator $I$. The degree function is defined corresponding to the regularity. For the space-time noise $\xi$ on $[0,T]\times D\in\mathbb{R}^d$, its Hölder regularity is $-\epsilon-(d+2)/2$ for any small $\epsilon>0$, so we define $\rm deg\xi=-(d+2)/2$. \citep{chevyrev}

To help understand the rule of generating models, we provide the elements of models in the three experiments. All the operators $I$ and $I_c$ below have been defined as the ones in the experiment section. And for the 2d equation, we use the $I_i, I_{c_i}$ to denote $\partial I/\partial x_i, \partial I_c/\partial x_i$ respectively, where $i=1,2$.

\begin{enumerate}
    \item  Dynamic $\phi^4_1$ model: We note that $\beta$ in (\ref{equa:deg}) is 2 according to the definition of $I$. As for $\alpha$, we take the forcing order $p=1$ because the forcing $\xi$ only appears once. As $\mu$ and $\sigma$ do not depend on $\partial_i$, we take the differentiation order $q=0$. We construct a model with additive width $m=3$, multiplicative width $l=1$, i.e. $\alpha=(3,1,1,0)$, and degree $\le7.5$.
    \begin{enumerate}
        \item $n=2: I[\xi], I_c[u_0], I[I[\xi]], I[I_c[u_0]], I[(I_c[u_0])^2], I[(I_c[u_0])(I[\xi])], I[(I[\xi])^2],\\
        I[(I_c[u_0])^2(I[\xi])], I[(I_c[u_0])(I[\xi])^2], I[(I[\xi])^3]$.
        \item $n=3: I[\xi], I_c[u_0], I[I[\xi]], I[I_c[u_0]], I[(I_c[u_0])^2], I[(I_c[u_0])(I[\xi])], I[(I[\xi])^2],\\
        I[(I_c[u_0])^2(I[\xi])], I[(I_c[u_0])(I[\xi])^2], I[(I[\xi])^3], I[I[I[\xi]]], I[I[I_c[u_0]]], I[I[(I_c[u_0])^2]],\\
        I[I[(I_c[u_0])(I[\xi])]], I[I[(I[\xi])^2]], I[I[(I_c[u_0])^2(I[\xi])]], I[I[(I_c[u_0])(I[\xi])^2]],\\
        I[I[(I[\xi])^3]], I[(I[I_c[u_0]])(I[\xi])], I[(I[(I_c[u_0])(I[\xi])^2])(I[\xi])], I[(I[I[\xi]])^2],\\
        I[(I[I[\xi]])(I[\xi])], I[(I[I[\xi]])(I_c[u_0])], I[(I[I[\xi]])(I[(I[\xi])^2])], I[(I[\xi])(I[(I[\xi])^3])],\\
        I[(I[\xi])(I_c[u_0])], I[(I[\xi])(I[(I_c[u_0])(I[\xi])])], I[(I[\xi])(I[(I[\xi])^2])],\\
        I[(I[(I[\xi])^3])(I_c[u_0])], I[(I_c[u_0])(I[(I[\xi])^2])], I[(I[I_c[u_0]])(I[\xi])^2], I[(I[I[\xi]])^2(I[\xi])],\\
        I[(I[I[\xi]])(I[\xi])^2], I[(I[I[\xi]])(I[\xi])(I_c[u_0])], I[(I[\xi])^2(I[(I[\xi])^3])], I[(I[\xi])^2(I_c[u_0])],\\
        I[(I[\xi])^2(I[(I_c[u_0])(I[\xi])])], I[(I[\xi])^2(I[(I[\xi])^2])], I[(I[\xi])(I_c[u_0])^2],\\
        I[(I[\xi])(I_c[u_0])(I[(I[\xi])^2])]$.
    \end{enumerate} 
    \item Reaction-Diffusion Equation with Linear Multiplicative Forcing: As the form of this equation is almost the same as the $\Phi^4_1$ model, the model $\mathcal{M}$ can be constructed similarly: the operator $I$, the initial index set $\mathcal{J}$ and $u^c$ are all same. What differs is the $\alpha$ because the change of the forcing term. Due to the multiplicative forcing, the multiplicative width $l=2$, i.e. $\alpha=(3,2,1,0)$. Only elements whose degrees do not exceed 7.5 are involved as well.
    \begin{enumerate}
        \item $n=2: I[\xi], I_c[u_0], I[I[\xi]], I[I_c[u_0]], I[\xi(I[\xi])],I[\xi(I_c[u_0])], I[(I[\xi])^2],\\
        I[(I[\xi])(I_c[u_0])],I[(I[\xi])^3], I[(I[\xi])^2(I_c[u_0])]$.
        \item$n=3: I[\xi], I_c[u_0], I[I[\xi]], I[I_c[u_0]], I[\xi(I[\xi])], I[\xi(I_c[u_0])],  I[(I[\xi])^2],\\
        I[(I_c[u_0])(I[\xi])], I[(I[\xi])^3], I[(I_c[u_0])(I[\xi])^2], I[I[I[\xi]]], I[I[I_c[u_0]]], I[I[\xi(I_c[u_0])]],\\
        I[I[\xi(I[\xi])]], I[I[(I_c[u_0])(I[\xi])]], I[I[(I[\xi])^2]],I[I[(I_c[u_0])(I[\xi])^2]], I[I[(I[\xi])^3]],\\
        I[\xi(I[\xi(I[\xi])])], I[\xi(I[(I[\xi])^3])], I[\xi(I[I_c[u_0]])], I[\xi(I[I[\xi]])], I[\xi(I[(I_c[u_0])(I[\xi])^2])],\\
        I[\xi(I[\xi(I_c[u_0])])], I[\xi(I[(I[\xi])^2])], I[\xi(I[(I_c[u_0])(I[\xi])])], I[(I_c[u_0])(I[\xi(I[\xi])])],\\
        I[(I[\xi(I[\xi])])^2],I[(I[\xi(I[\xi])])(I[\xi])], I[(I[I[\xi]])(I[\xi])], I[(I[\xi])(I[\xi(I_c[u_0])])],\\
        I[(I[\xi(I[\xi])])^3], I[(I[\xi(I[\xi])])^2(I[\xi])], I[(I[\xi(I[\xi])])(I[\xi])^2]$.
    \end{enumerate}
    \item 2d stochastic Navier-Stokes equation: We construct the model $\mathcal{M}$ with $\alpha=(2,1,1,1)$, $\rm deg\le7.5$. As for the $\alpha$, we note that the right side of the NSE contains $\nabla$, so the differentiation order is set to $q=1$.
    \begin{enumerate}
        \item $n=2: I[\xi], I_c[\omega_0], I[I[\xi]], I[I_1[\xi]], I[I_2[\xi]], I[I_c[\omega_0]], I[I_{c_1}[\omega_0]], I[I_{c_2}[\omega_0]],\\
        I[(I_c[\omega_0])^2], I[(I_c[\omega_0])(I_{c_2}[\omega_0])], I[(I_c[\omega_0])(I_{c_1}[\omega_0])], I[(I_c[\omega_0])(I_2[\xi])],\\
        I[(I_c[\omega_0])(I[\xi])], I[(I_c[\omega_0])(I_1[\xi])], I[(I_{c_2}[\omega_0])^2], I[(I_{c_2}[\omega_0])(I_{c_1}[\omega_0])],\\
        I[(I_{c_2}[\omega_0])(I_2[\xi])], I[(I_{c_2}[\omega_0])(I[\xi])], I[(I_{c_2}[\omega_0])(I_1[\xi])], I[(I_{c_1}[\omega_0])^2],\\
        I[(I_{c_1}[\omega_0])(I_2[\xi])], I[(I_{c_1}[\omega_0])(I[\xi])], I[(I_{c_1}[\omega_0])(I_1[\xi])], I[(I_2[\xi])^2], I[(I_2[\xi])(I[\xi])],\\
        I[(I_2[\xi])(I_1[\xi])], I[(I[\xi])^2], I[(I[\xi])(I_1[\xi])], I[(I_1[\xi])^2]$.
        \item $n=3: I[\xi], I_c[\omega_0], I[I[\xi]], I[I_1[\xi]], I[I_2[\xi]],I[I_c[\omega_0]], I[I_{c_1}[\omega_0]], I[I_{c_2}[\omega_0]],\\
        I[(I_c[\omega_0])^2],I[(I_c[\omega_0])(I_{c_1}[\omega_0])],
       ...,
       I[(I[(I_1[\xi])(I_2[\xi])])(I[(I_1[\xi])(I[\xi])])],\\ I[(I[(I_1[\xi])(I_2[\xi])])(I[(I_1[\xi])(I_{c_2}[\omega_0])])],
       I[(I[(I_1[\xi])(I_2[\xi])])(I[(I_{c_1}[\omega_0])(I[\xi])])],\\
       I[(I[(I_c[\omega_0])(I[\xi])])(I[(I_1[\xi])(I[\xi])])],
       I[(I[(I_{c_1}[\omega_0])^2])(I[(I_1[\xi])(I[\xi])])],\\
       I[(I[(I_1[\xi])(I[\xi])])^2],
       I[(I[(I_1[\xi])(I[\xi])])(I[(I_1[i])(I_{c_2}[\omega_0])])],\\
       I[(I[(I_1[\xi])(I[\xi])])(I[(I_{c_1}[\omega_0])(I[\xi])])],
       I[(I[(I_1[\xi])(I_{c_2}[\omega_0])])^2],\\
       I[(I[(I_1[\xi])(I_{c_2}[\omega_0])])(I[(I_{c_1}[\omega_0])(I[\xi])])]$.
    \end{enumerate}
\end{enumerate}

\subsection{Additive Description about the Experiments}
For the code of NSPDE and FNO, please refer to \url{https://github.com/crispitagorico/Neural-SPDEs} and \url{https://github.com/zongyi-li/fourier_neural_operator}, respectively.

Apart from the FNO and NSPDE, we also compare NORS with other models on the first two experiments. The complete results are report in Table \ref{table:phi41_} and Table \ref{table:multi_}, from which we can see the superiority of NORS.

\begin{table}[h]
\caption{\textbf{Dynamic $\Phi^4_1$ model}. We consider the $l2$ error of the baselines and our model($\alpha=(3,1,1,0),\rm deg\le7.5$, $n=2$ or $3$) in two settings.}
\label{table:phi41_}
\begin{center}
\begin{tabular}{lcc|cc}
\multirow{2}{*}{\bf Model} & \multicolumn{2}{c}{$N=1000$} & \multicolumn{2}{c}{$N=10000$} \\
\cline{2-5}
& $\xi\mapsto u$ & $(u_0,\xi)\mapsto u$ & $\xi\mapsto u$ & $(u_0,\xi)\mapsto u$
\\ \hline \\
NCDE     & 0.112 & 0.127 & 0.056 & 0.072 \\
NRDE     & 0.129 & 0.150 & 0.070 & 0.083 \\
NCDE-FNO & 0.071 & 0.066 & 0.066 & 0.069 \\
DeepONet & 0.126 & $\times$ & 0.061 & $\times$ \\
FNO      & 0.032 & 0.030 & 0.027 & 0.024 \\
NSPDE    & 0.009 & 0.012 & 0.006 & 0.006 \\
\cline{1-5}
Ours($n=3$) & \textbf{0.0003} & \textbf{0.0011} & \textbf{0.0002} & \textbf{0.0004} \\
Ours($n=2$) & \textbf{0.0002} & \textbf{0.0012} & \textbf{0.0002} & \textbf{0.0004} \\
\end{tabular}
\end{center}
\end{table}

\begin{table}[h]
\caption{\textbf{Reaction-Diffusion Equation with Linear Multiplicative Forcing}. We compare the $l2$ error of the baselines and our model($\alpha=(3,2,1,0),\rm deg\le7.5$, $n=2$ or $3$) in two settings.}
\label{table:multi_}
\begin{center}
\begin{tabular}{lcc|cc}
\multirow{2}{*}{\bf Model} & \multicolumn{2}{c}{\bf $N=1000$} & \multicolumn{2}{c}{\bf $N=10000$} \\
\cline{2-5}
& $\xi\mapsto u$ & $(u_0,\xi)\mapsto u$ & $\xi\mapsto u$ & $(u_0,\xi)\mapsto u$
\\ \hline \\
NCDE     & 0.016 & 0.087 & 0.010 & 0.059 \\
NRDE     & 0.023 & 0.584 & 0.023 & 0.641 \\
NCDE-FNO & 0.015 & 0.034 & 0.017 & 0.019 \\
DeepONet & 0.023 & $\times$ & 0.023 & $\times$ \\
FNO      & 0.0036 & 0.0063 & 0.0035 & 0.0037 \\
NSPDE    & 0.0016 & 0.0062 & 0.0012 & 0.0026 \\
\cline{1-5}
Ours($n=3$) & \textbf{0.0006} & \textbf{0.0005} & \textbf{0.0005} & \textbf{0.0003} \\
Ours($n=2$) & \textbf{0.0006} & \textbf{0.0006} & \textbf{0.0005} & \textbf{0.0003} \\
\end{tabular}
\end{center}
\end{table}




\end{document}